\documentclass{sig-alternate}

\PassOptionsToPackage{pdfpagelabels=true}{hyperref}
\usepackage{todonotes}
\presetkeys{todonotes}{inline,backgroundcolor=yellow}{}

\usepackage{flushend}

\begin{document}

\setcopyright{rightsretained}

\doi{}

\isbn{}


\title{Exploring epoch-dependent stochastic residual networks}

\numberofauthors{1}
\author{
%
%
\alignauthor
Fabio Carrara, Andrea Esuli, Fabrizio Falchi, Alejandro Moreo Fern\'andez\\
       \affaddr{ISTI-CNR}\\
       \affaddr{via G. Moruzzi, 1}\\
       \affaddr{56124 Pisa, Italy}\\
       \email{\texttt{firstname.lastname}@isti.cnr.it}
}

\maketitle
\begin{abstract}
The recently proposed stochastic residual networks selectively activate or bypass the layers during training, based on independent stochastic choices, each of which following a probability distribution that is fixed in advance.
In this paper we present a first exploration on the use of an epoch-dependent distribution, starting with a higher probability of bypassing deeper layers and then activating them more frequently as training progresses.
Preliminary results are mixed, yet they show some potential of adding an epoch-dependent management of distributions, worth of further investigation.
\end{abstract}

\printccsdesc


\section{Introduction}
Residual networks have recently emerged as an effective device to exploit very deep neural networks \cite{he2015deep}.
\cite{huang2016deep} proposed to add a stochastic distribution to residual layers as a device to gain in efficiency while also reducing overfitting at training time.
In this latter sense, stochastic residual networks model a dropout-like effect along the network depth dimension.
In \cite{huang2016deep}, each layer was assigned a probability of being activated or bypassed during training; the probability of being active decreases linearly with the depth of the layer in the net.
The probability distribution was fixed before training and never changed during the learning process in \cite{huang2016deep}.

In this paper we explore the use of epoch-dependent stochastic distributions on residual networks.
Specifically, we use a stochastic distribution that activates more rarely the deeper layers of the network on the initial epochs of training, and then increases the probability of such activation as the training progresses. 
This can be viewed also as using a less deep network, in terms of expected depth, at the early stages of learning and then slightly increasing its depth after each epoch.
The rationale of this idea is to have a network that is less prone to overfit, and/or to be unstable, on the deepest levels, i.e., those that should capture the 
most abstract characteristics of the data,
at the early stages of learning, when the first levels of the network have not yet stabilized.
As learning progresses, deeper layers are more frequently activated, allowing the network to model higher-level information from the training data.

We test various configurations and report results on the CIFAR-10 and CIFAR-100 datasets \cite{krizhevsky2009learning}, showing that the use of an epoch-depended stochastic distribution may have a positive impact on error rate.

\section{Growing Res-Nets}
Given a ResNet composed by $L$ residual blocks, \cite{huang2016deep} assigns the survival
probability of the $l^{th}$ residual block $p_l$ following a fixed linear decay:
$p_l$ linearly decreases from $1$ to $p_L$, where $p_L$ is a hyper-parameter that determines the survival probability of the last residual block.
This can be expressed using death rates $d_l = 1 - p_l$ as follows:
$$	
d_l=\frac{l}{L} \cdot d_L
$$

During the forward-backward pass, dead residual blocks are skipped and not trained, resulting in a shallower network with a stochastic number of residual blocks $\tilde{L} < L$. 
The expected number of residual blocks is given by
$$
E(\tilde{L}) = \sum_{l=1}^{L} p_l = L \cdot \left ( 1 - \frac{d_L}{2} \right )
$$

We define a \textit{growing} ResNet as a stochastic residual network where all $p_l$ vary (increase) during the training process.
Given $k \in [0,1]$ the current training epoch number normalized by the maximum number of epochs, let $d_l^k$ denote the death rate of the $l$-th residual block at epoch $k$. 
We set
\begin{eqnarray} 
d_L^k &=& \left ( (1-k) \cdot d_L^0 + k \cdot d_L^1 \right ) \\
d_l^k &=& \frac{l}{L} d_L^k \label{eq:grow}
\end{eqnarray}
where $d_L^0$ and $d_L^1$ are new hyper-parameters of the model that determine respectively the initial and final death rate for the last block. 
With this assignment of $d_l^k$, the expected length at each epoch $\tilde{L^k}$ is given by:
$$
E(\tilde{L^k}) = L \cdot \left ( 1 - \frac{d_L^k}{2} \right )
$$

We also tested a \emph{linear growth} model in which a value $\omega$ determines the depth, at the first epoch, at which blocks have death rate one.
Then the death rates have a linear evolution following the equations:
\begin{equation}  \label{eq:aggr-lgrow}
d_l^k = \max\left (0,\min \left ( 1, \frac{l-k}{\omega} \right )\right)
\end{equation}

We finally experimented a more \emph{aggressive growth} model, in which only some of the first residual blocks are active at the beginning of the training phase, i.e., the deeper blocks have $d_l^k=1$.
To implement this growth mode, we set
\begin{equation} \label{eq:aggr-grow}
d_l^k = \min \left ( 1, (1-k) \cdot \frac{l}{L \cdot s} \right )
\end{equation}
where $s$ is the percentage of residual blocks which can be active at the beginning of the training.

These last two models can be seen as an incremental fine-tuning of several networks increasing in depth, each of which is initialized with weights pretrained at the previous epoch. 

\begin{figure*}
	\begin{minipage}[t]{.5\linewidth}
		\includegraphics[width=\linewidth]{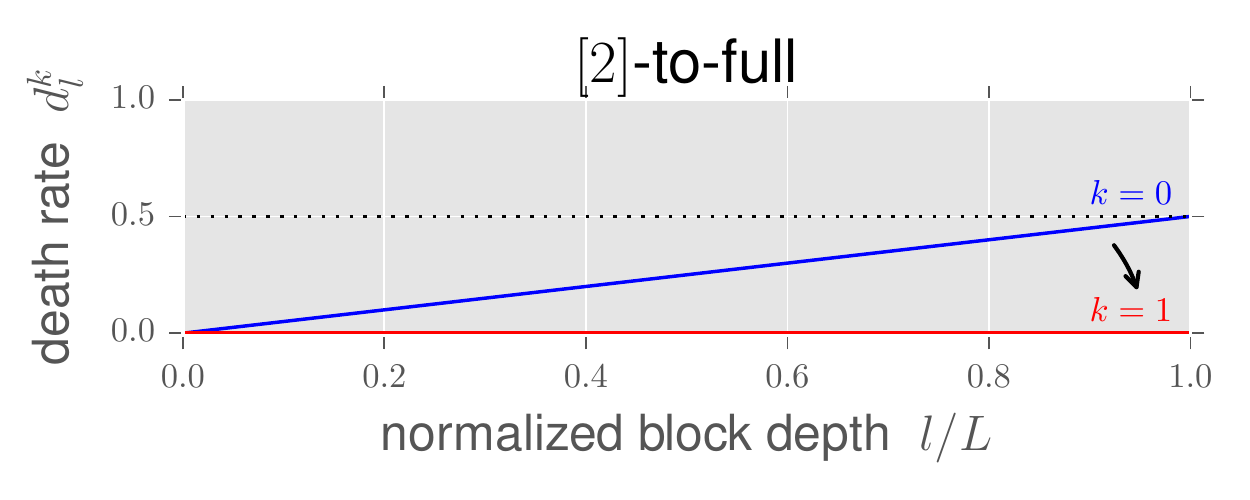}
  		\includegraphics[width=\linewidth]{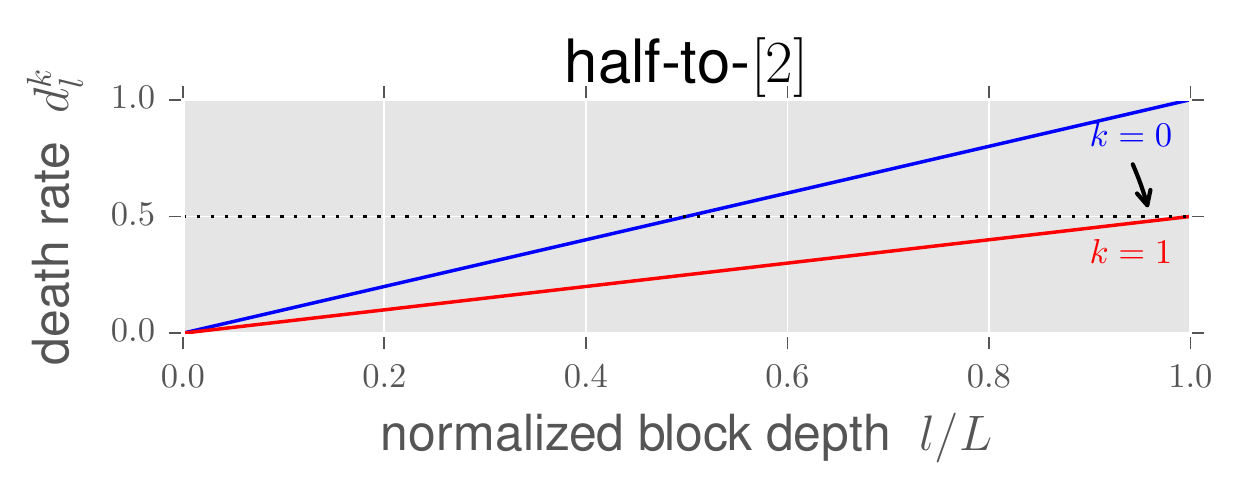}
        \includegraphics[width=\linewidth]{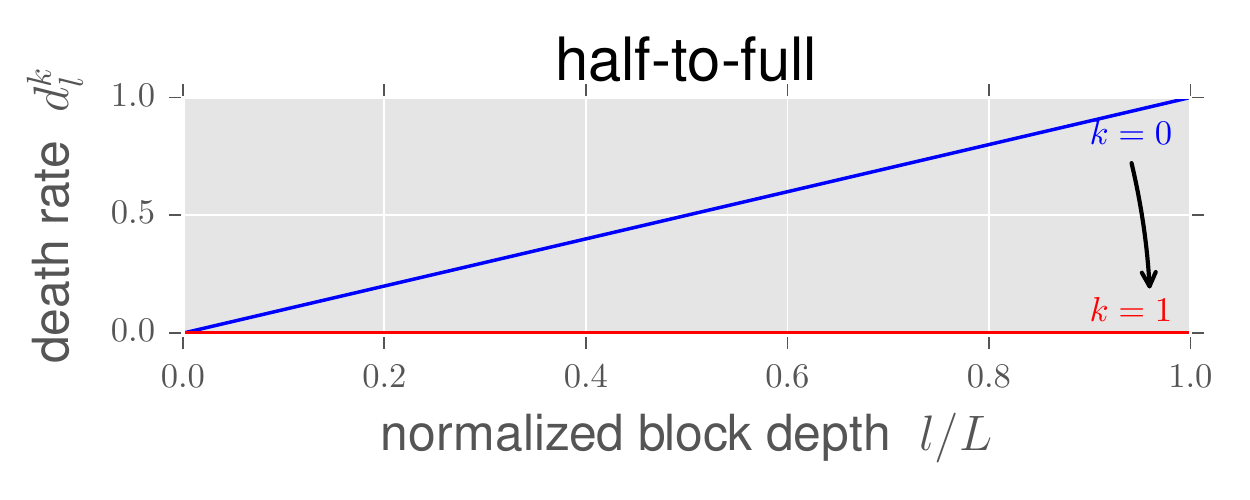}
        \vspace{10em}
        
		\includegraphics[width=\linewidth]{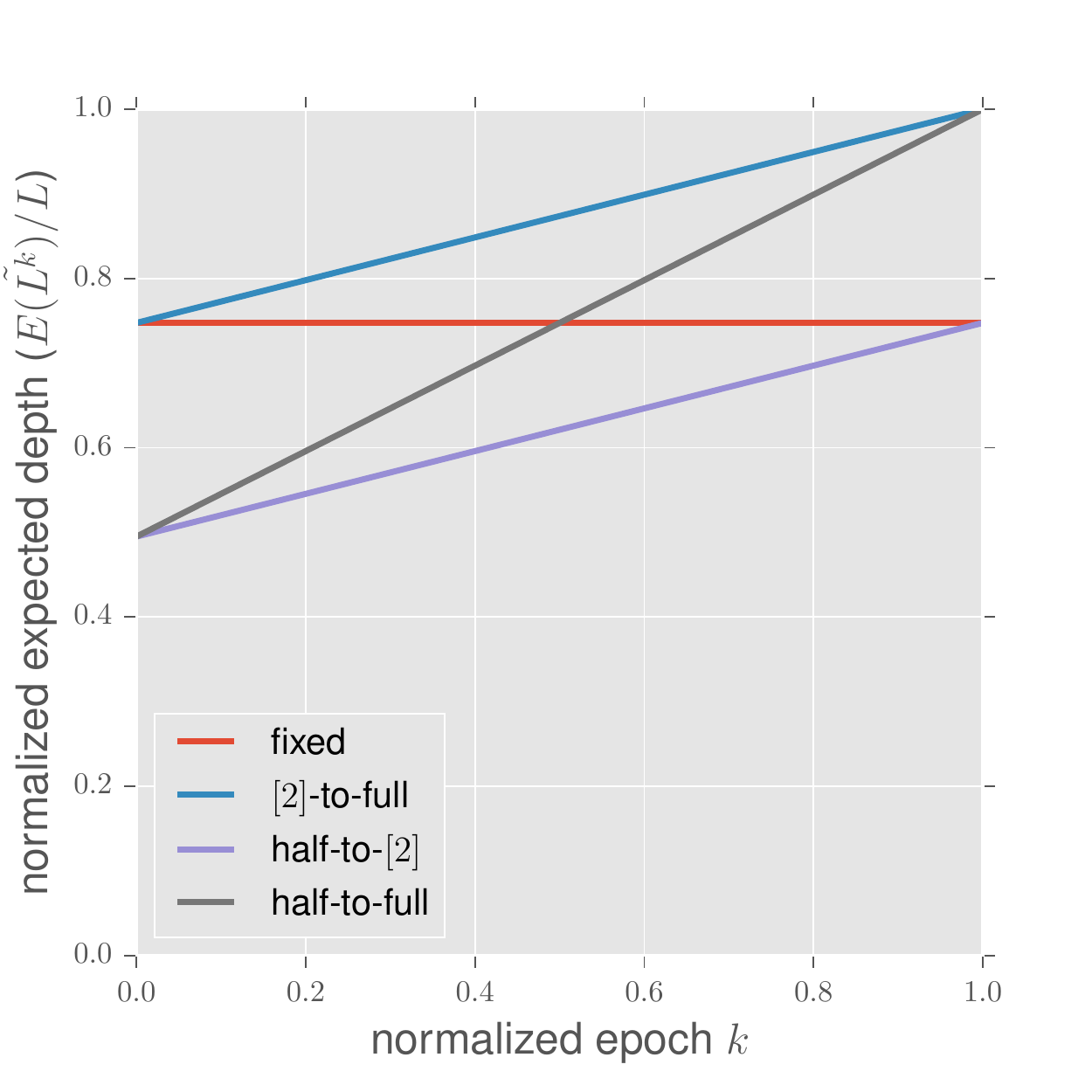}
	\end{minipage}
    \begin{minipage}[t]{.5\linewidth}
        \includegraphics[width=\linewidth]{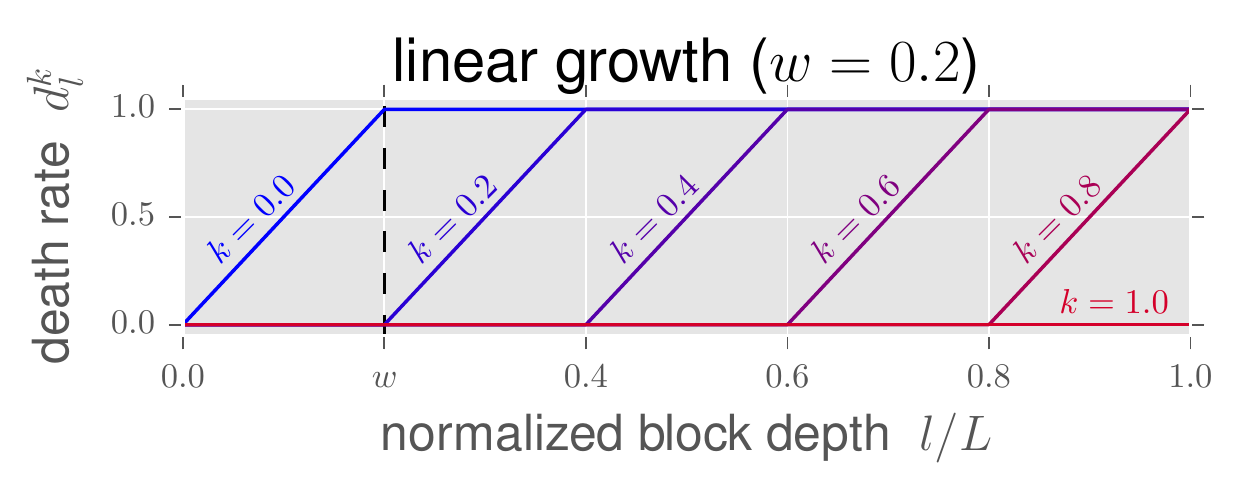}
        \includegraphics[width=\linewidth]{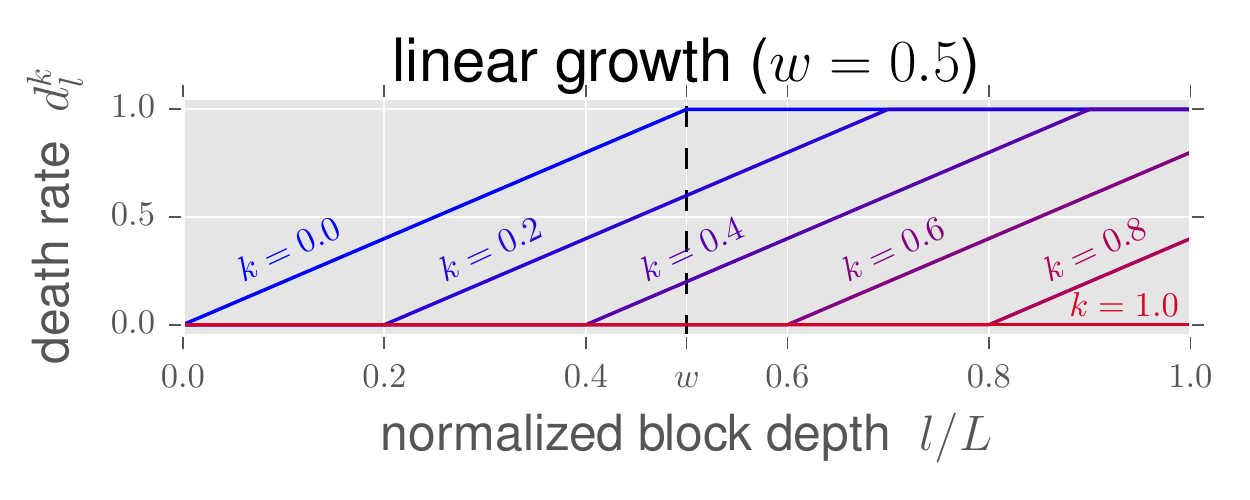}
        \includegraphics[width=\linewidth]{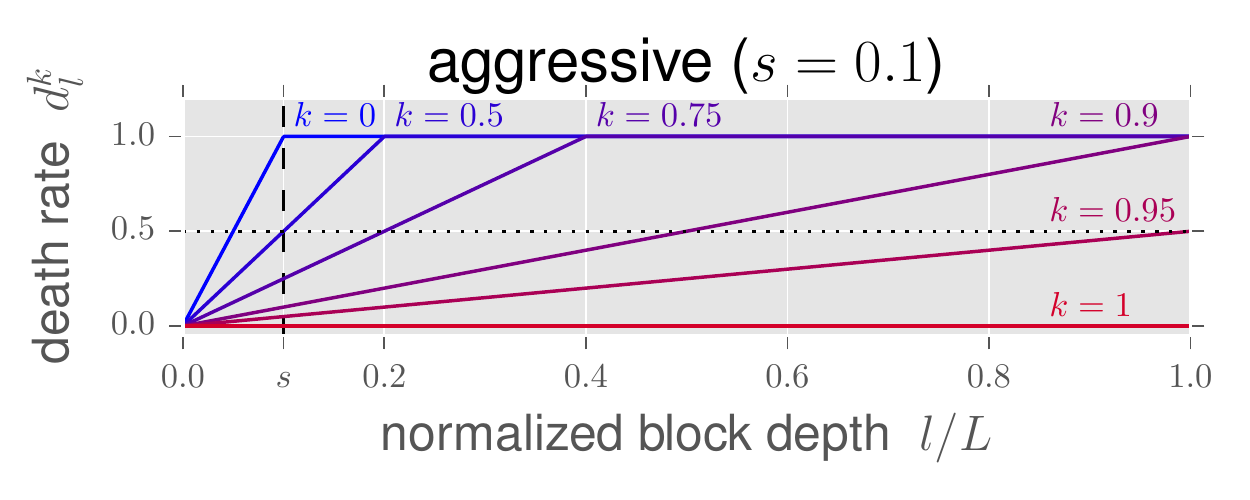}
        \includegraphics[width=\linewidth]{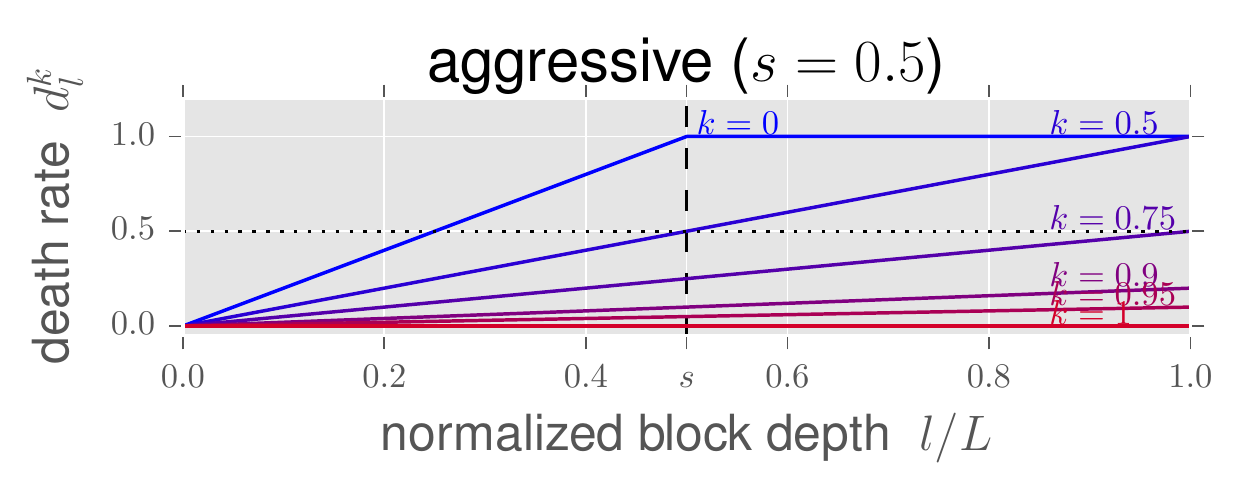}
   		\includegraphics[width=\linewidth]{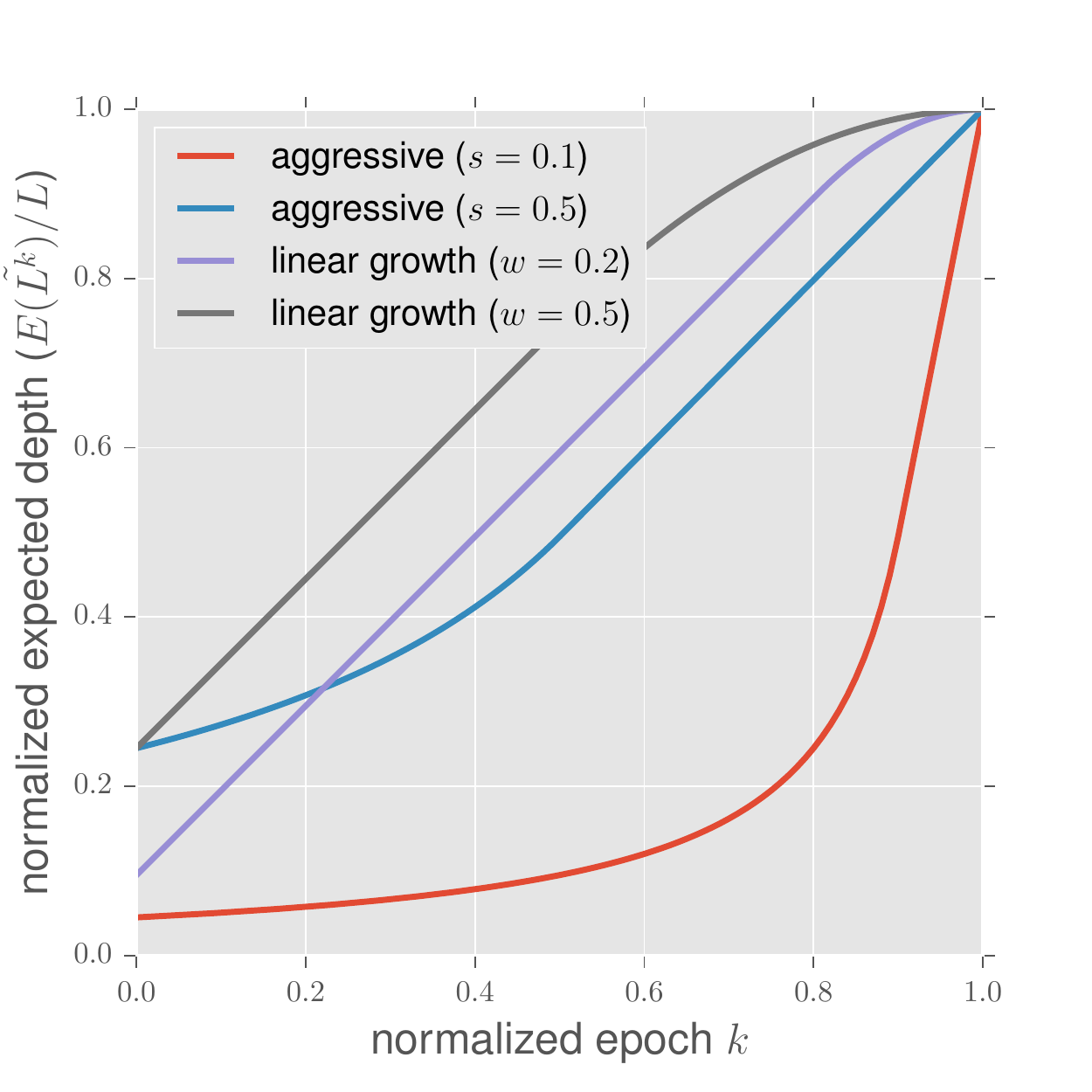}
	\end{minipage}
    \label{fig:growth-modes}
	\caption{Evolution of death rates and expected network depth in experimented configurations.
    Top graphs show the death rates of residual blocks as a function of the normalized position in the network $l/L$ and the current normalized epoch number $k$. Bottom graphs show the normalized expected network depth as a function of $k$}
\end{figure*}

\section{Experiments}
\subsection{Datasets}

Among many available datasets for image classification, we choose to perform our experiments on CIFAR-10 and CIFAR-100 \cite{krizhevsky2009learning}. 
Those datasets are composed by tiny images, and allow us to train several computational intense models, such as very deep residual networks, in a reasonable amount of time.

CIFAR-10 is composed by 60.000 color images with a 32x32 resolution, divided in 10 classes, each composed by 6.000 images. 
CIFAR-100 is also composed by 60.000 32x32 images, belonging to 100 different classes, each composed by 600 images. 
Training and test subsets are already provided, and they are composed respectively by 50.000 and 10.000 images for both datasets.

For each dataset, we held out 5.000 images of the training subset for validation, and we applied common data augmentation techniques (random horizontal flipping and translation) to the remaining images in order to match the setup of \cite{huang2016deep}.

\subsection{Training}

We trained several 110-layer ResNets on both CIFAR-10 and CIFAR-100 datasets with different configurations of $p_l$.
First, we built a model with an initial assignment of $p_l$ that matches the configuration used in \cite{huang2016deep} (dubbed \emph{fixed}), following a linear decay from $p_1 = 1$ to $p_L = 0.5$.

We compared three growth models based on the Equation~\ref{eq:grow}.
In one setup we decreased all the death probabilities at each epoch until all of them reached $0$  at the end of the training phase (dubbed \emph{\cite{huang2016deep}-to-full}). 
This is achieved setting $d_l^k$ as in Equation~\ref{eq:grow} with $d_L^0 = 0.5$ and $d_L^1 = 0$ (Figure~\ref{fig:growth-modes}, first row on the left).

The second setup sets $d_L^0 = 1$ and $d_L^1 = 0.5$, with which the model starts with the deepest layer inactive ($p_l = 0$), and ends its training phase with the configuration of $p_l$ proposed in \cite{huang2016deep} (dubbed \emph{half-to-\cite{huang2016deep}}, Figure~\ref{fig:growth-modes}, second row on the left). 
In this configuration, we grow the residual network from an initial stochastic depth of $L / 2$ to a final expected depth of $3L / 4$.

The third setup explores a wider variation, starting with the same configuration of \emph{half-to-\cite{huang2016deep}} and concluding with the same configuration of \emph{\cite{huang2016deep}-to-full}, i.e., $d_L^0 = 1$ and $d_L^1 = 1$.
We dub this configuration \emph{half-to-full}, note that the average expected depth of this configuration across all epochs is the same of \cite{huang2016deep}. 

On the \emph{linear growth} model of Equation \ref{eq:aggr-lgrow} we tested two values of $\omega=0.5$ and $\omega=0.2$.
Finally, we trained a number of aggressive growth modes following the Equation~\ref{eq:aggr-grow}. 
In a very aggressive setup we used $s = 0.1$, meaning that we initially train a shallow network with a maximum depth of $s \cdot L$, and we grow it until it reaches the full depth at the end of the training phase.
We also tested a relatively less aggressive setup with a value $s=0.5$.
The non-linear evolution of the expected network depth thus imposed is visualized in Figure~\ref{fig:growth-modes}, in the lower-right plot.

We trained all the models with Nesterov's accelerated gradient for 500 epochs, with initial learning rate of 0.1, divided by 10 at epochs 250 and 375, as reported in \cite{huang2016deep}.

\subsection{Results}
\begin{figure*}[t!]
\begin{center}
\includegraphics[width=\linewidth]{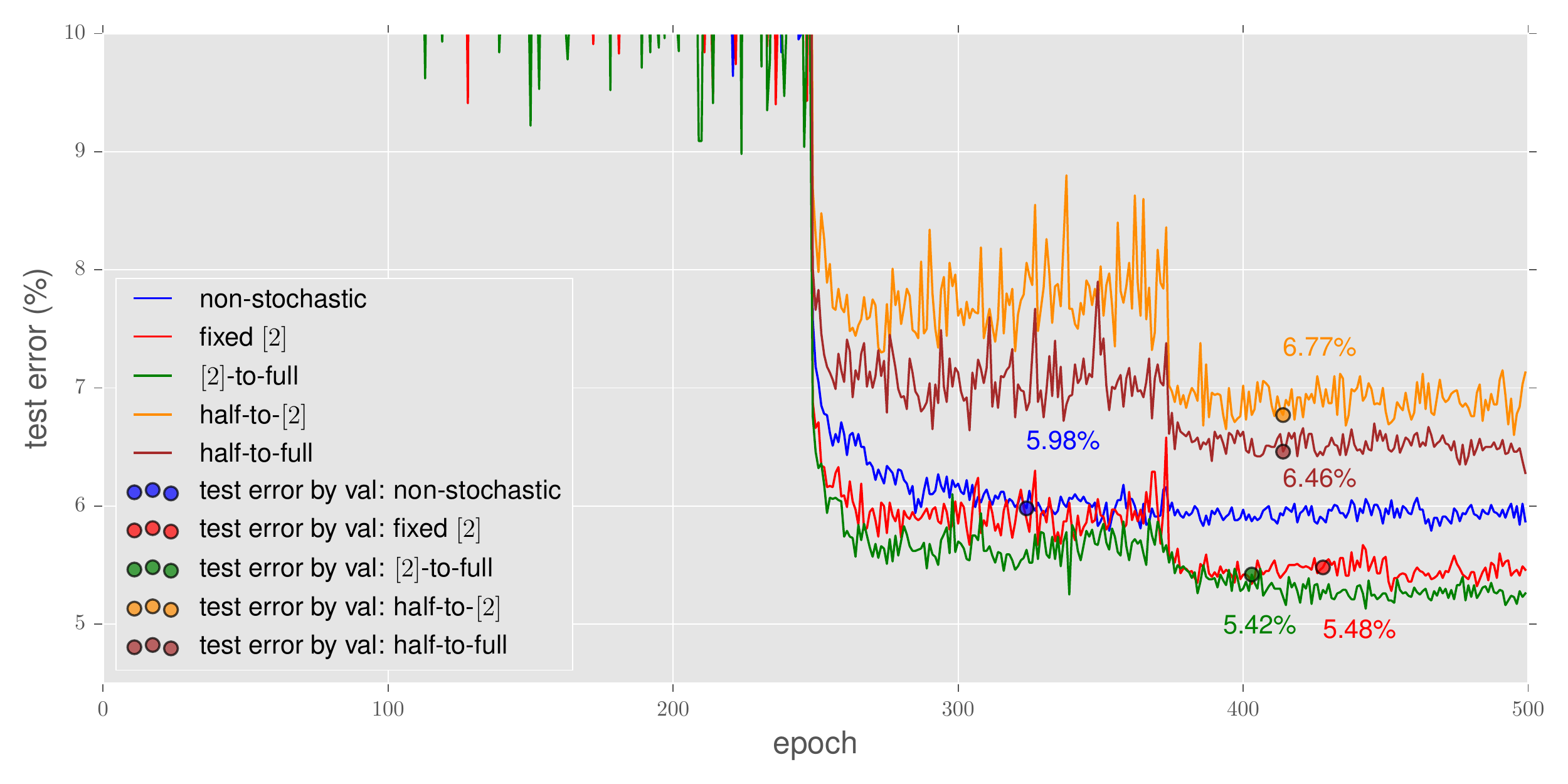}
\includegraphics[width=\linewidth]{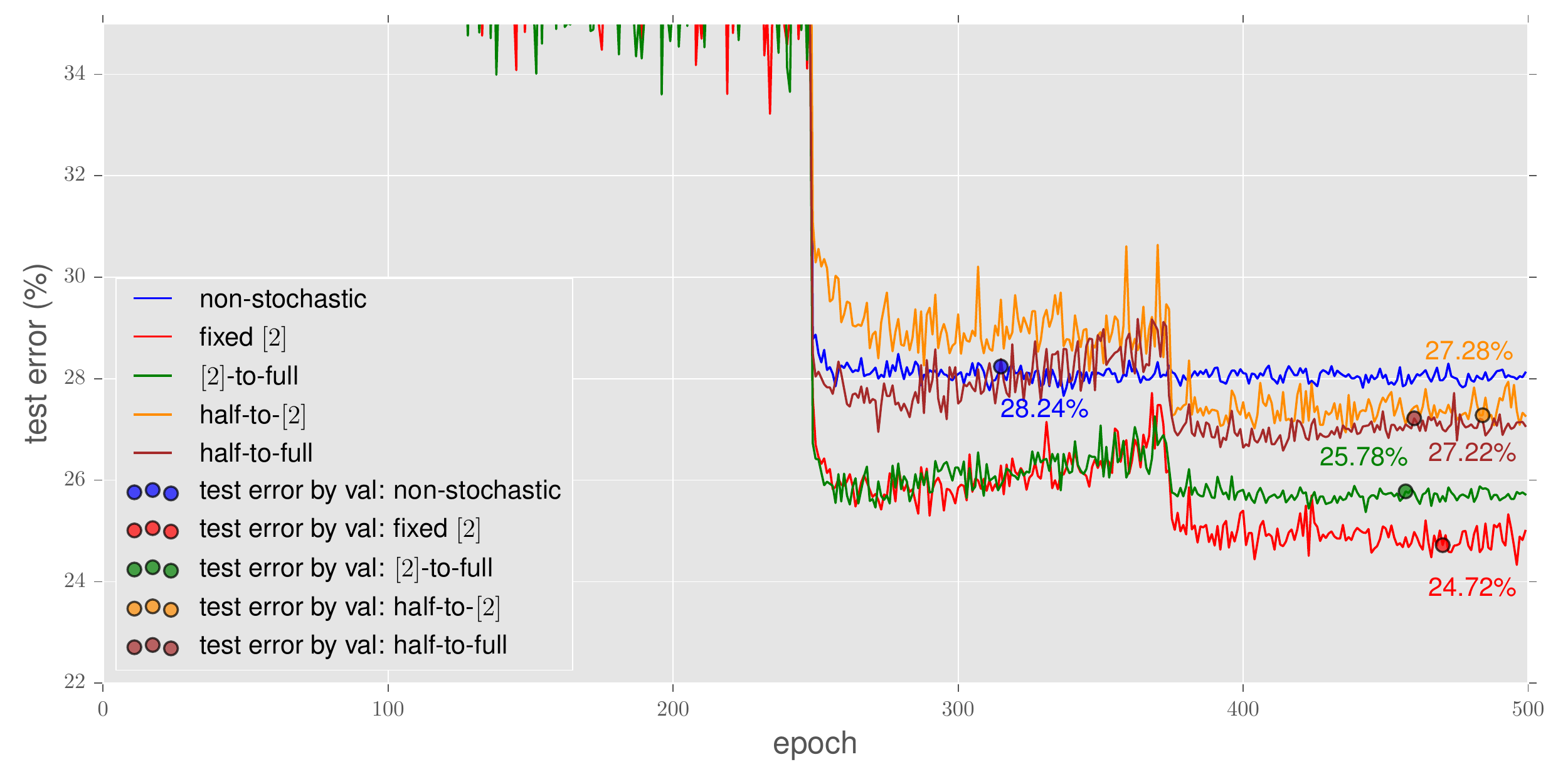}%
\end{center}
\label{fig:results10}
\vspace{-5ex}\caption{Test error on CIFAR-10 (top) and CIFAR-100 (botton) during training. Dots indicate the test error selected by validation.}
\end{figure*}

Figure \ref{fig:results10} show the validation error during training and the resulting test error, including a \emph{normal} setup that simply uses the full network, without any stochastic filtering.

The \emph{\cite{huang2016deep}-to-full} setup obtains a  marginal, yet significant, improvement over the \emph{fixed} setup on CIFAR-10, but it is worst on CIFAR-100.
The worst performance of \emph{\cite{huang2016deep}-to-full} with respect to \emph{fixed} on CIFAR-100 is not predictable until the last change of learning rate.
Our hypothesis is that, being the \emph{\cite{huang2016deep}-to-full} a deeper network than \emph{fixed}, the learning rate change should be posticipated. 
We leave this to future investigation.

The \emph{half-to-\cite{huang2016deep}} configuration performed worst than the non-stochatic network on CIFAR-10 and better than it on CIFAR-10, however always worst than the other two similar configurations. 
In favor of the \emph{half-to-\cite{huang2016deep}} configuration there is the fact that it is sensibly shorter than the full net (60\% on average).

The experiments of \emph{linear growth} and \emph{aggressive} models performed sensibly worse than the others (out of the scale of the plots).
The worst results is for the \emph{aggressive} model with $s=0.1$ that obtained a test error of 21.87\% on CIFAR-10 and 59.55\% on CIFAR-100.
This setup has proved to be too much aggressive, leaving on average too few active layers in a large part of the epochs. 
In fact, the average length of the network is only approximately $19\%$ of the total length $L$.

\section{Conclusions}
All the aggressive growth modes performed poorly in our experiments, we deem this poor performance to an excess of aggressiveness on the expected depth, as shown in Figure \ref{fig:growth-modes}. 
E.g., at 80\% of the total epochs only 20\% of the full depth is used.
The \emph{\cite{huang2016deep}-to-full} growth mode obtained promising results.
We observe that the interaction between the learning rate and the variation of the expected depth of the network in an interesting dimension to investigate, not only changing the epochs at which the learning rate is changed, but also experimenting with other optimization methods, e.g., Adam \cite{adam}.



%

%
%

\end{document}